\documentclass[10pt,journal,compsoc]{IEEEtran}
\ifCLASSOPTIONcompsoc
  \usepackage[nocompress]{cite}
\else
  \usepackage{cite}
\fi
\ifCLASSINFOpdf
\else
\fi
\usepackage{graphicx}
\usepackage{amsmath}
\usepackage{url}

\begin{document}
%
\title{Observer Design for Augmented Reality-based Teleoperation of Soft Robots}
%
%
%
%

\author{Jorge~Francisco García-Samartín,
        Iago~López~Pérez,
        Emirhan~Yolcu,
        Jaime~del~Cerro
        and~Antonio~Barrientos
\IEEEcompsocitemizethanks{\IEEEcompsocthanksitem J.F.G, I.L.P, J.C. and A.B are with the Centro de Automática y Robótica, Universidad Politécnica de Madrid, c/ José Gutiérrez Abascal, nº 2, 28006, Madrid, España.\protect\\
E-mail: jorge.gsamartin@upm.es
\IEEEcompsocthanksitem E.Y. is with Middle East Technical University, Dumlupinar Blv No 1, 06800 Cankaya, Ankara, Türkiye.}
\thanks{~}}

%
%

\markboth{Observer Design for Augmented Reality-based Teleoperation of Soft Robots}%
{García-Samartín, J. F., Lórez Pérez, I., Yolcu, E., del Cerro, J., Barrientos, A.}
%



\IEEEtitleabstractindextext{%
\begin{abstract}
Although virtual and augmented reality are gaining traction as teleoperation tools for various types of robots, including manipulators and mobile robots, they are not being used for soft robots. The inherent difficulties of modelling soft robots mean that combining accurate and computationally efficient representations is very challenging. This paper presents an augmented reality interface for teleoperating these devices. The developed system consists of Microsoft HoloLens 2 glasses and a central computer responsible for calculations. Validation is performed on PETER, a highly modular pneumatic manipulator. Using data collected from sensors, the computer estimates the robot's position based on the physics of the virtual reality programme. Errors obtained are on the order of 5\% of the robot's length, demonstrating that augmented reality facilitates operator interaction with soft manipulators and can be integrated into the control loop.
\end{abstract}

}

\maketitle

\IEEEdisplaynontitleabstractindextext

%
\IEEEpeerreviewmaketitle

\section{Introduction}
In recent decades, numerous virtual (VR) or augmented reality (AR) interfaces for remotely operating robotic systems have been developed for diverse applications such as surgery~\cite{Wang2024AR}, search and rescue~\cite{CruzUlloa2024Analysis}, industrial manipulation~\cite{Gong2025Augmented}, repairing tasks~\cite{Aschenbrenner2019} and rehabilitation~\cite{Makhataeva2020Augmented}. This is largely due to the advantages they offer operators, such as more intuitive control, greater feedback, reduced cognitive load and enhanced safety, particularly in hazardous environments~\cite{Naceri2021The, Su2023Integrating}.

However, its applications in soft robotics are still very limited. Current simulation options in virtual reality software, coupled with the limitations of existing computing power, still prevent the faithful real-time reconstruction of the pose of these manipulators. This is because the information provided by the sensors is often indirect, requiring an observer to relate their readings to the robot's state.

The first successful implementations of augmented or virtual reality in robots of this type relate to hyperredundant robots~\cite{Martin-Barrio2020b}. Soft robots can be considered an extreme example of this, with infinite modules. Other examples include mixed systems, such as the rehabilitation system presented in~\cite{HE2025104994}.

\begin{figure*}[t]
\centering
  \includegraphics[width=\linewidth]{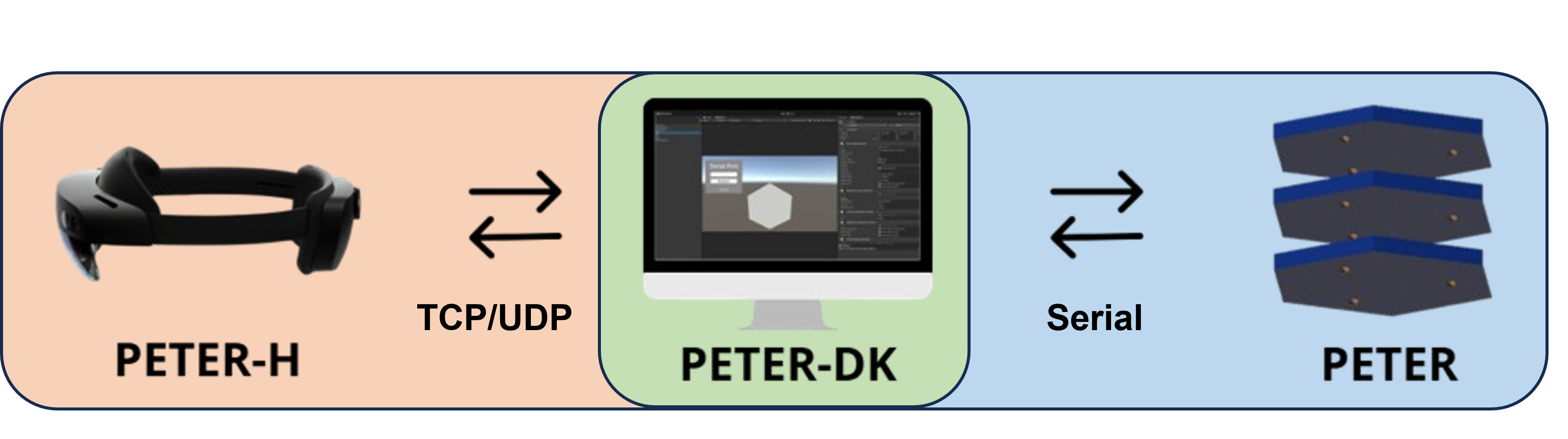}\\
  \caption{System Architecture.}
  \label{fig:mat-arch}
\end{figure*}

Three works stand out as particularly relevant in the field of soft robotics itself. Firstly, \cite{Eslami2023} uses VR as an interface for controlling a silicone pneumatic finger. As it is a single-degree-of-freedom system with a curvature sensor, the reconstruction is straightforward and assumes constant curvature despite potential disturbances.

The same reconstruction philosophy is employed in~\cite{Borges2022} to model a Pneunet-type actuator in AR. However, greater accuracy is achieved by using six coloured markers as sensors, whose shape is reconstructed based on vertex transforms.

Finally, \cite{Bern2024} uses a FEM-based algorithm to observe the state of a multi-degree-of-freedom cabled robot that is teleoperated with mixed reality. While the accuracy achieved is remarkable, the movement times are slow, raising the question of how much precision is necessary. Furthermore, the developed simulator appears to be highly specific to the robot in question, offering limited customisation options. Additionally, it does not address any aspects related to user experience in remote operation.

This work presents a new augmented reality system for teleoperating a modular, parallel, six-input pneumatic platform. The system uses two inertial measurement units (IMUs) and two time-of-flight (TOF) sensors for measurement. These sensors send information to the augmented reality system via a central computer, which then reconstructs the robot's pose. As in existing literature, therefore, the augmented reality system acts as the observer of the robot's state. Special attention is given to the design of the user interface, and extensive options for robot customisation are provided.

Thus, it can be said that the work's two main contributions are as follows:
\begin{itemize}
    \item Firstly, the development of an augmented reality interface for teleoperating modular soft robots. Although some interfaces exist, they are very limited, and aspects such as customisation have not previously been addressed. This article presents a complete and intuitive interface.
    \item Secondly, the use of Unity as a state observer is explored. Although its precision will always be lower than that of FEM or NN-based solutions, this integrated solution has great potential since a virtual model of the robot must always be created and its accuracy evaluated.
\end{itemize}

The rest of the article is structured as follows: Section~\ref{secc:mat} describes the proposed system, which comprises the physical manipulator, the AR interface and the Unity model. Section~\ref{secc:obs} presents the state observer. Section~\ref{secc:obs} describes the results, and Section~\ref{secc:concl} draws the conclusions.

\section{Augmented Reality System}
\label{secc:mat}

\subsection{System Overview}

The developed AR system, whose architecture is depicted in Figure~\ref{fig:mat-arch}, aims to teleoperate pneumatic soft robots composed of modules with various actuators arranged in a parallel structure. To validate the system, the PETER manipulator, which was presented in previous work and consists of two modules, was used. Microsoft HoloLens 2~\cite{Microsoft2026} headsets were employed and an interface called PETER-H was developed for these. The system model runs on a central computer and is called PETER-DK.

PETER-DK is responsible for receiving sensor information from the real robot, transmitting control commands back to it and indicating to the interface which model to display. PETER-DK communicates with the physical manipulator via a serial connection, while the central computer communicates with the headset via the TCP/UDP protocol.

\subsection{PETER Manipulator}
\begin{figure}[!ht]
\centering
  \includegraphics[width=0.8\linewidth]{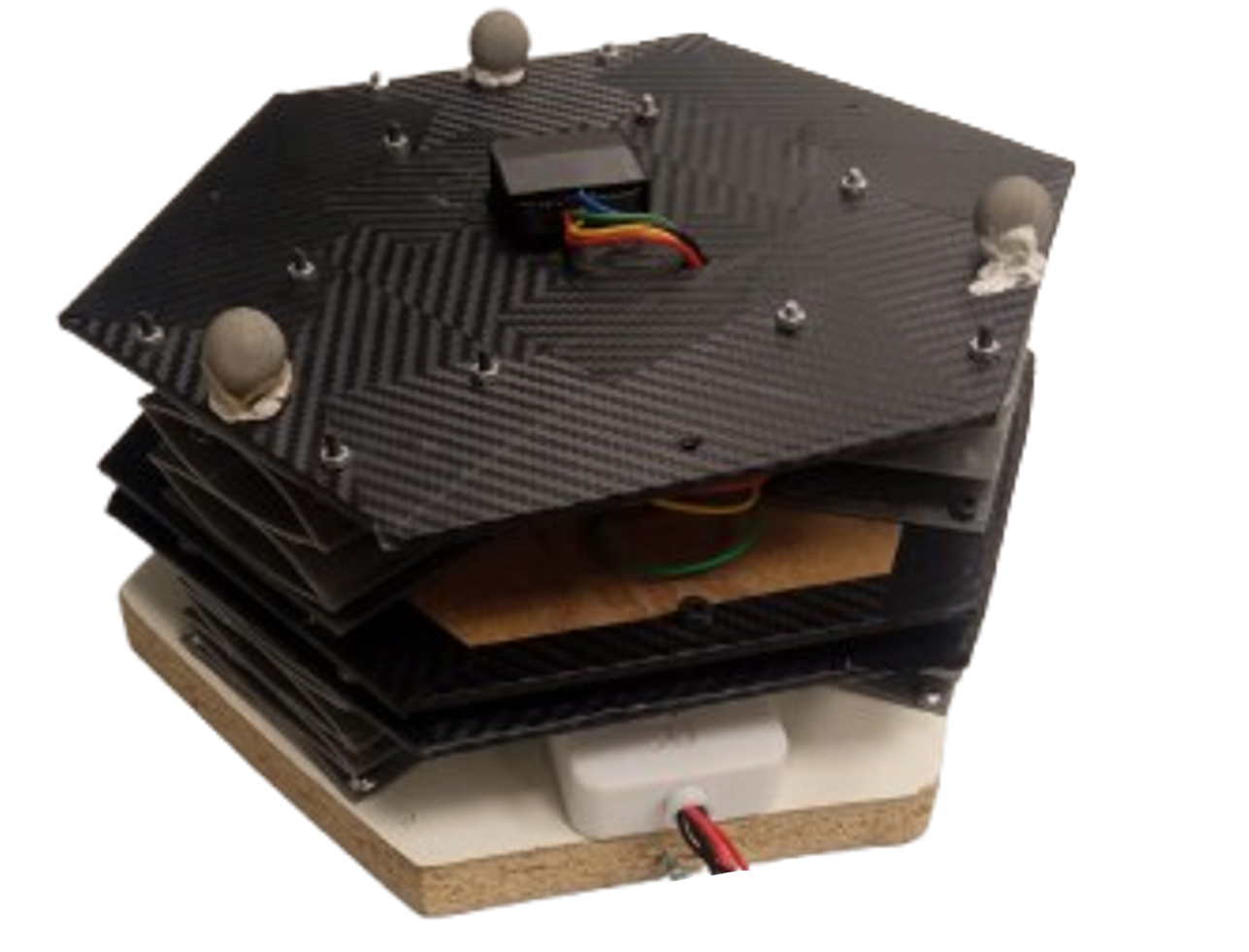}\\
  \caption{PETER Manipulator.}
  \label{fig:mat-peter}
\end{figure}

Figure~\ref{fig:mat-peter} shows PETER, a pneumatic manipulator. Consisting of two modules, each of which is composed of three parallel actuators, it has five degrees of freedom and six air inlets. Each actuator has a TPU structure and inflatable rubber components, providing high load capacity and extensibility. Each module incorporates a TOF and an IMU, enabling unique sensorisation. Construction details can be found in~\cite{Garcia-Samartin2024d}.

Previous work has addressed the resolution of its kinematics~\cite{Garcia-Samartin2025}. However, it is difficult for operators who are not specialised in robotics to understand it properly, as well as to manage obstacles in the environment. This makes the development of interfaces like the one presented in this work highly advantageous.

\subsection{PETER-H Interface}
\begin{figure}[!ht]
\centering
  \includegraphics[width=\linewidth]{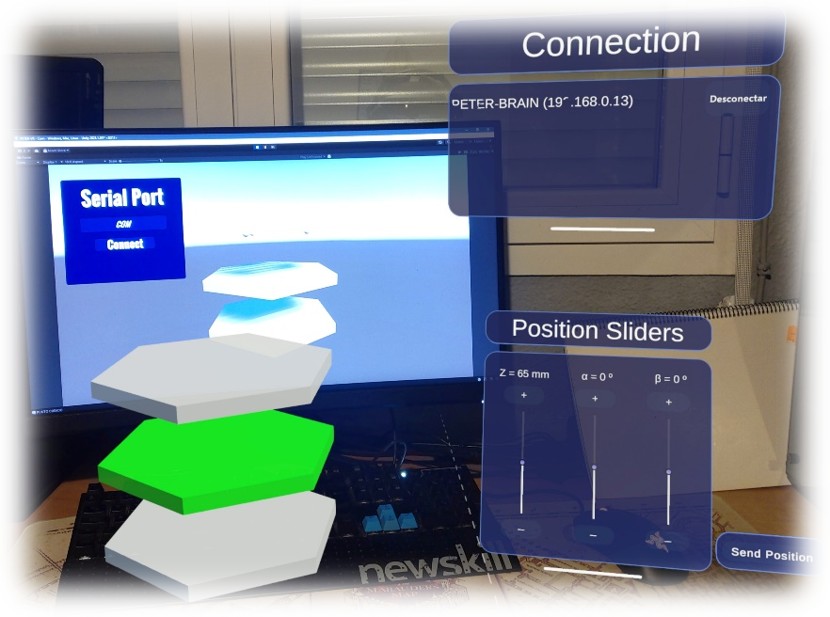}\\
  \caption{PETER-H Interface with PETER-DK on the back.}
  \label{fig:mat-int}
\end{figure}

\begin{figure}[!ht]
\centering
  \includegraphics[width=\linewidth]{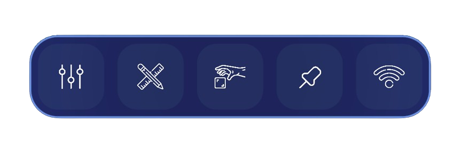}\\
  \caption{PETER-H Menu. The available options are: toggle control sliders, edit geometric parameters, drag the PETER model, fix the PETER model and connect to PETER-DK.}
  \label{fig:mat-menu}
\end{figure}

The second element of the system is the AR interface, which is represented in Figure~\ref{fig:mat-int}. The central element of this is the robot's visualisation. Alongside this are sliders for setting the desired configuration and buttons for indicating that PETER should be moved to that position. There is also a menu, depicted in Figure~\ref{fig:mat-menu}., for managing the connection with the central computer and customising the manipulator's geometric parameters.

\begin{figure}[!ht]
\centering
  \includegraphics[width=\linewidth]{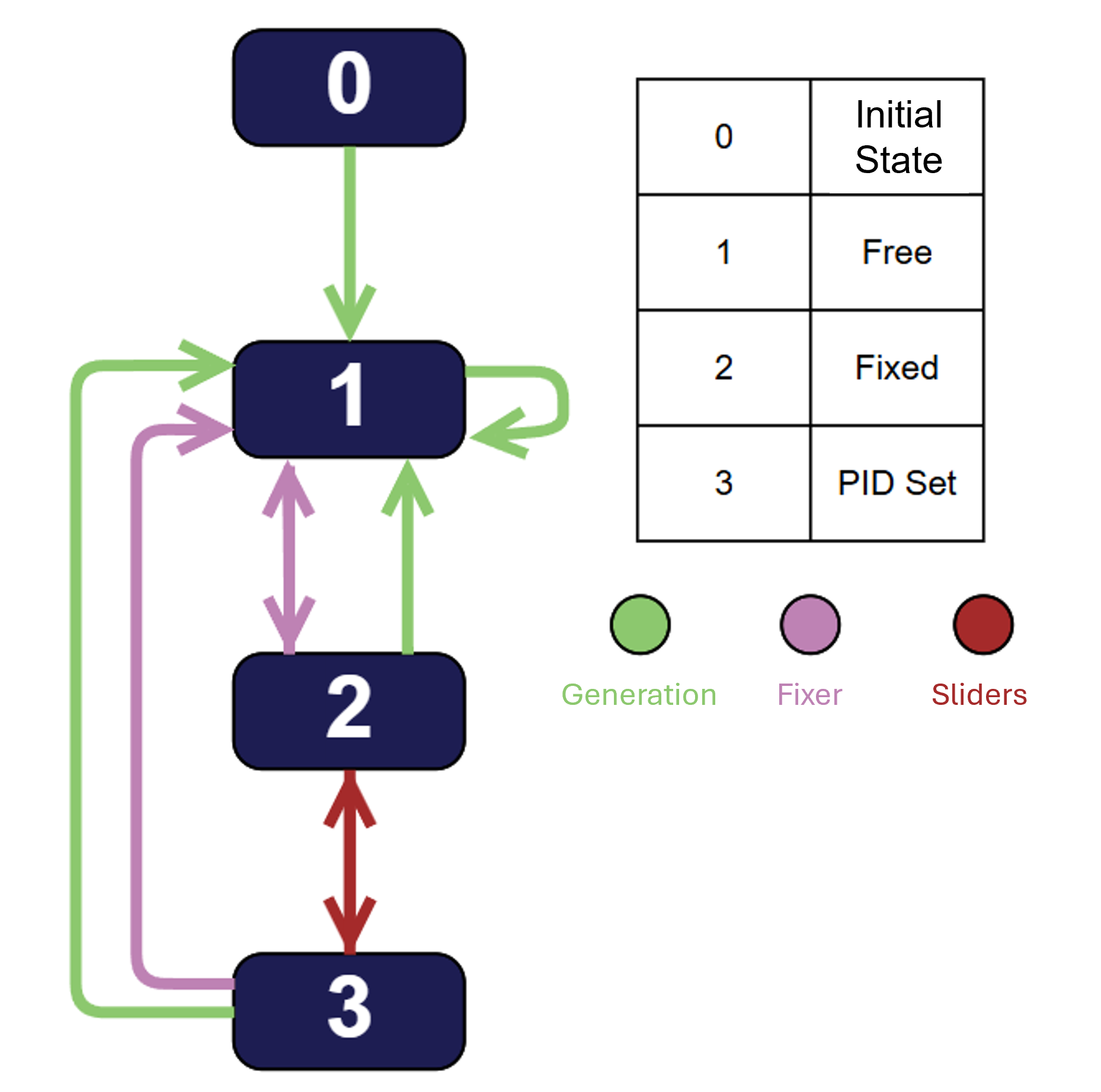}\\
  \caption{PETER-H State Diagram.}
  \label{fig:mat-states}
\end{figure}

This subsystem operates sequentially, as described in the state diagram of Figure~\ref{fig:mat-states}. Once connected to PETER-DK, the interface enables users to configure the robot's various geometric options, as detailed in Section~\ref{subsecc:mat-dk}. A virtual robot is then generated, passing from State 0 to State 1 that the user can drag and rotate freely within their field of view until the lock button is pressed, in which system enters in State 2. One of the platforms can then be selected, and its position adjusted using the sliders. At this point, the operator can ask the physical PETER to move to the specified position and enter State 3. A PID will then be activated until PETER reaches its position with an error of 3 mm or less.

\subsection{PETER-DK Server}
\label{subsecc:mat-dk}

This subsystem is responsible for receiving and processing sensor data for visualisation and control purposes, and for providing PETER-H with the necessary visualisation instructions.

PETER-DK was developed entirely in Unity. One of the objectives in the design phase was to make it suitable for the broadest range of parallel soft robots possible. To achieve this, the module was established as the basic unit of calculation, meaning the system estimates the kinematics of each module sequentially before concatenating them.

The user can define various parameters for each module, including the number of actuators, the radius of the polygon whose vertices are the actuators, the platform height, and the maximum and minimum elongation of the actuators. These parameters are calculated to ensure safety and prevent the physical manipulator from being sent with excessively inflated actuators that could cause leaks or breakage.

\begin{figure}[!ht]
\centering
  \includegraphics[width=\linewidth]{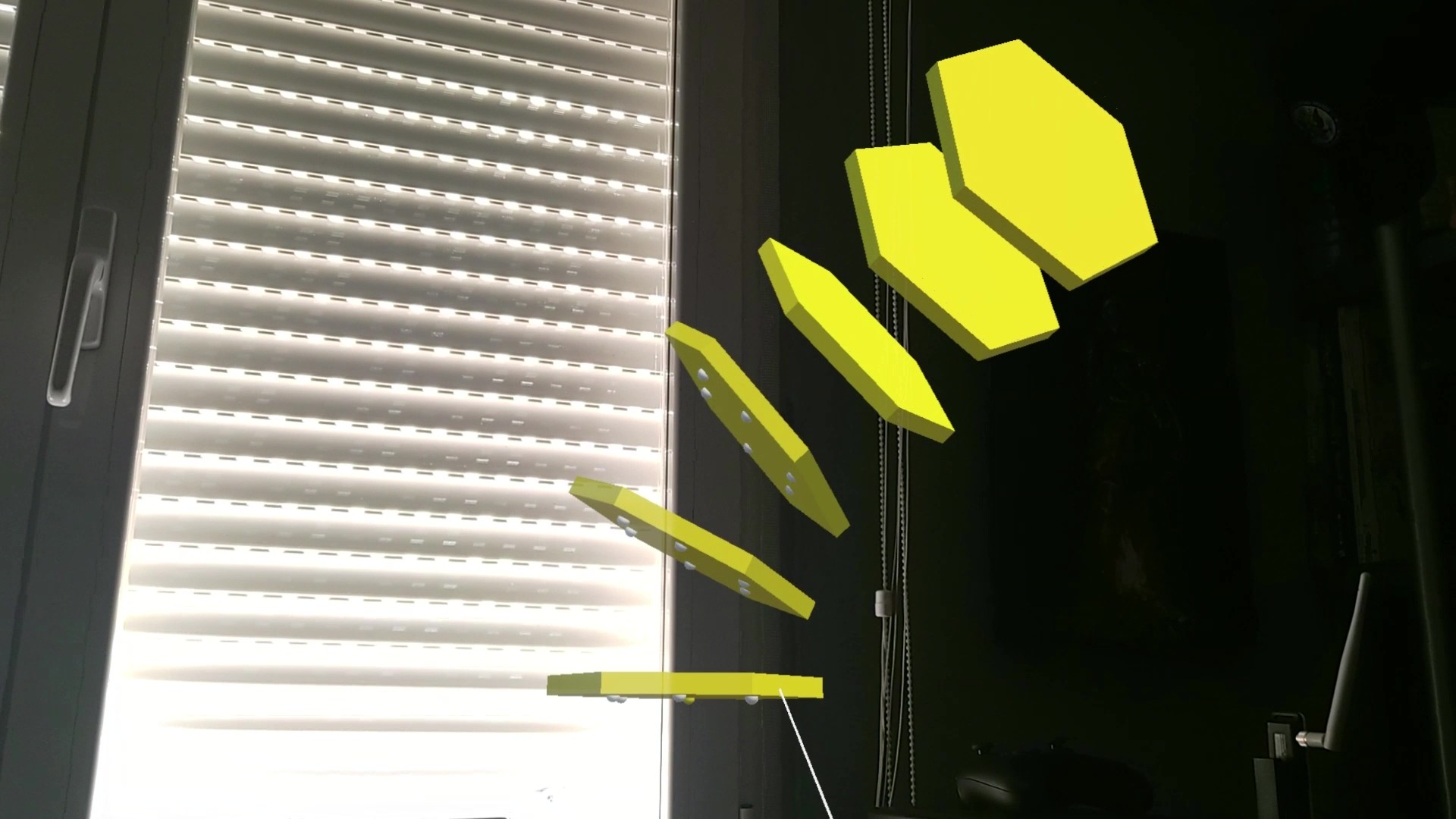}\\
  \caption{AR representation of PETER with 5 modules, each with 3 actuators.}
  \label{fig:mat-mult}
\end{figure}

Figure~\ref{fig:mat-mult} shows an example of a virtual robot with multiple modules. PETER-DK has been demonstrated to calculate the kinematics of 9 modules in real time, showcasing its capabilities. However, this does not mean that the system as a whole can function with such a large number of modules, as managing communications would likely result in high latency.

\section{Observer Design}
\label{secc:obs}

\begin{figure}[!ht]
\centering
  \includegraphics[width=0.8\linewidth]{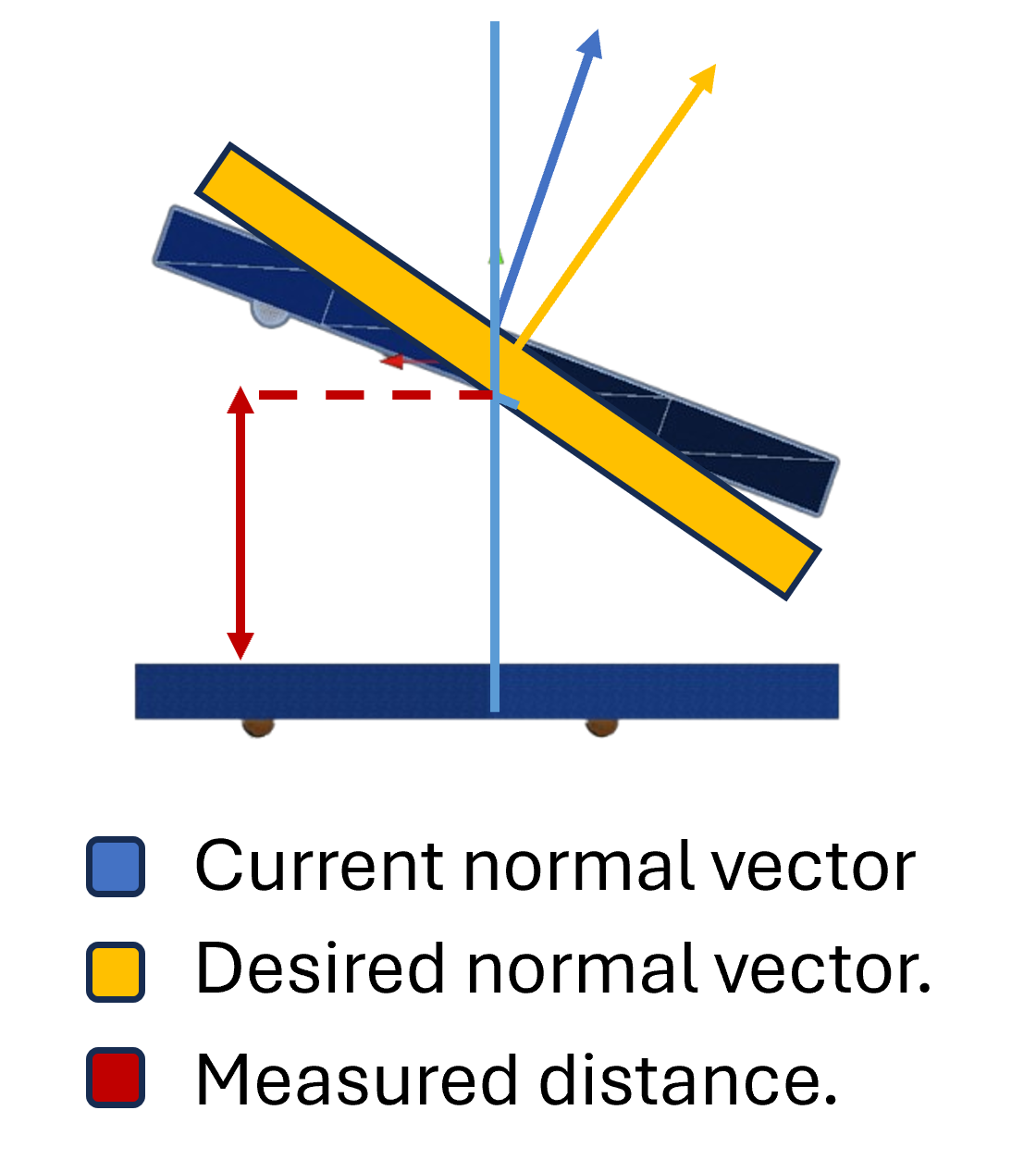}\\
  \caption{Model assumptions: rotation over a fixed point and height equal to the average height of the actuators.}
  \label{fig:obs-diagr}
\end{figure}

In addition to designing an AR-based teleoperation interface for soft robots, this work aims to study the interface's models and their ability to predict the state of soft robots. Therefore, when configuring the model, it is essential to establish geometric and physical hypotheses that are consistent with reality but also simple enough to allow for real-time execution.

From the outset, therefore, the pneumatic nature of the system was disregarded and the actuators were considered as extendable rods. The inflation commands issued by PETER-DK to the physical manipulator correspond to the lengths calculated by the central computer, as characterised in~\cite{Garcia-Samartin2025}. This simplification, based on good results obtained in previous work, significantly reduced computational cost with no noticeable loss of precision.

Unity's physics determined the lengths necessary for the central point of the module to reach a point $(x, y, z)$. These configurations were based on the assumption that the upper platform of each module rotated around its fixed central point and that this central point was always at an average height equal to that of the actuators. Unlike the usual assumption in robotics of constant curvature, this hypothesis simplifies the calculations and is valid because the length of the actuators is similar to their separation, and no module reaches curvatures greater than 10°. Figure~\ref{fig:obs-diagr} shows a representative diagram of these assumptions.

For each module, PETER-DK receives the height value measured by the TOF sensor, $h$, as well as the IMU's rotation values around the X and Y axes, denoted by $\varphi$ and $\vartheta$, respectively. The program processes these values and informs Unity of the lengths of the actuators $l_{1}, l_{2}$ and $l_{3}$.

Assuming each module has 3 actuators forming a triangle with side $L$, the positions of its bases are the ones specified in Equation~\eqref{eq:obs-bases}:
\begin{align}
    \boldsymbol{a}_0 &= [L, 0, 0]^T, \notag\\
    \boldsymbol{b}_0 &= \left[-\frac{L}{2}, \frac{\sqrt{3}L}{2}, 0\right]^T, \notag\\
    \boldsymbol{c}_0 &= \left[-\frac{L}{2}, -\frac{\sqrt{3}L}{2}, 0\right]^T
    \label{eq:obs-bases}
\end{align}

Rotation matrix of module $i$ with respect to module $i-1$ can be obtained directly from the readings of the IMU by considering a rotation $\varphi$ around the X-axis and a rotation $\vartheta$ around the Y-axis, as illustrated in Equation~\eqref{eq:obs-rotmat}:
\begin{equation}
    {}^{i-1}\boldsymbol{R}_{i} = \boldsymbol{R}_x(\varphi)\boldsymbol{R}_y(\vartheta) = \begin{bmatrix}
                                \cos \varphi & \sin \varphi \sin \vartheta & \sin \varphi \cos \vartheta \\
                                0 & \cos \vartheta & -\sin \vartheta \\
                                -\sin \varphi & \cos \varphi \sin \vartheta & \cos r \cos \vartheta
                            \end{bmatrix}
    \label{eq:obs-rotmat}
\end{equation}

The position of the end of each actuator on the upper platform is then calculated by applying the rotation matrix to its lower end and adding a displacement on the Z axis,, as shown in Equation~\eqref{eq:obs-top}:
\begin{align}
    \boldsymbol{a}_{i} = {}^{i-1}\boldsymbol{R}_{i}\boldsymbol{a}_{i-1} + [0, 0, h]^T\notag\\
    \boldsymbol{b}_{i} = {}^{i-1}\boldsymbol{R}_{i}\boldsymbol{b}_{i-1} + [0, 0, h]^T\notag\\
    \boldsymbol{c}_{i} = {}^{i-1}\boldsymbol{R}_{i}\boldsymbol{c}_{i-1} + [0, 0, h]^T
    \label{eq:obs-top}
\end{align}

PETER-DK uses this information to calculate the length of the actuators $\boldsymbol{l}_{i}$, the position of the centre of the upper platform, $\boldsymbol{x}_{i}$, as well as its normal vector, $\boldsymbol{n}$. This parameter is necessary in order to determine the position of the centre of the lower platform of the next module, $\boldsymbol{x}^{base}_{i+1}$. As the upper platform is rigidly connected to the lower one, with their centres separated by a distance d, the calculation is performed using the Equation~\eqref{eq:obs-next}:
\begin{equation}
    \boldsymbol{x}^{base}_{i+1} = \boldsymbol{x}_{i} + d\,\boldsymbol{n}'
    \label{eq:obs-next}
\end{equation}




If Rotation matrix of module $i$ in global axis, ${}^{0}\boldsymbol{R}_{i}$ is desired, it can be obtained multiplying the matrix in local axis by all the previous rotation matrix, as Equation~\eqref{eq:obs-rotglb} presents:
\begin{equation}
    {}^{0}\boldsymbol{R}_{i} =  {}^{0}\boldsymbol{R}_{1} {}^{1}\boldsymbol{R}_{2} \cdots {}^{i-1}\boldsymbol{R}_{i}
    \label{eq:obs-rotglb}
\end{equation}

Although the IMU measurements were relatively stable, the TOF exhibited significant noise in the form of sharp peaks that occurred randomly every few samples. To combat this, it was decided to filter the measurements. Although a mean filter seemed the most obvious choice, it was found that outliers excessively altered the result unless a very large number of measurements were used. This resulted in a delay that was too high for the control requirements. Therefore, a Kalman filter was fitted experimentally. Values of $Q = 0.01$ and $R = 0.40$ were used, which indicated much greater confidence in the model than in the TOF measurements.

\section{Results}
\label{secc:res}

\begin{figure*}[t]
\centering
  \includegraphics[width=\linewidth]{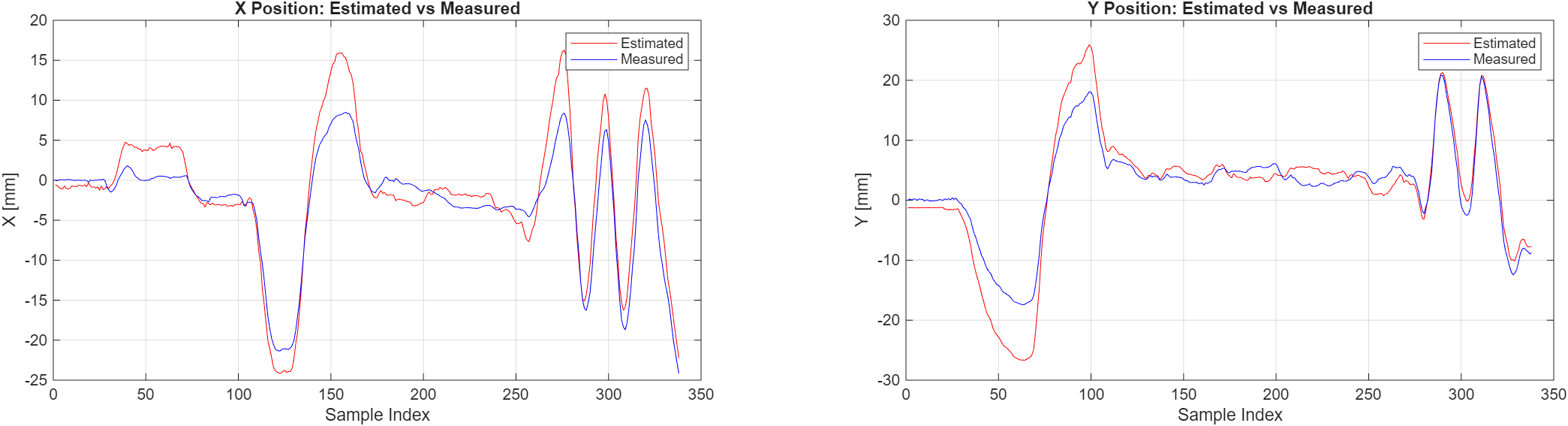}\\
  \caption{Comparison between the observed position (in red) and the real position (in blue) along the X and Y axis. Dsitances are displayed in mm.}
  \label{fig:res-xy}
\end{figure*}

\begin{table*}[t]
    \centering
    \caption{Error Statistical Analysis. Errors are displayed in mm.}
    \label{tab:errors}
    \begin{tabular}{lrrrrrr}
        \hline
        Metric & MAE & RSME & Std. Dev. & Max. MAE & Q1 (MAE) & Q3 (MAE) \\ \hline
        X axis & 2.41 & 3.04 & 2.78 & 7.87 & 0.98 & 3.61 \\ 
        Y axis & 2.59 & 3.56 & 3.55 & 9.35 & 1.09 & 2.96 \\ 
        Global & 3.93 & 4.68 & 2.54 & 10.19 & 2.04 & 5.37 \\ \hline
    \end{tabular}
\end{table*}

Finally, the performance of the various subsystems and the complete system was evaluated. The AR interface and its characteristics of usability, adaptability and comfort have not been studied in depth and remain a future area of research. The development of a complete extended reality system for soft robots has been considered a commendable achievement in itself, given the scarcity of existing solutions.

However, to verify these systems' validity as state observers, the error between the position estimates provided by PETER-DK and those achieved by PETER in reality was analysed. A 15-camera OptiTrack system with an accuracy of around one-tenth of a millimetre was used as the reference standard.

During the test, the robot followed a trajectory for one minute. Every 100 ms, the observer described in Section~\ref{secc:obs} estimated the robot's position, while the Optitrack system captured the actual position. A total of 339 samples were obtained and distributed throughout the robot's workspace. Figure~\ref{fig:res-xy} shows the trajectories along the X and Y axes.

As depicted, the trajectory estimated by the robot is very similar to the actual one, particularly at the central points. When the robot reaches the far ends of its workspace, the system tends to overestimate the robot's position.

The same phenomenon can be observed in Table~\ref{tab:errors}, where the difference between MAE and RMSE is approximately 0.7 mm. This indicates the presence of large point errors of up to 10 mm, which significantly increase the overall error. However, considering that the combined length of the two modules is 85 mm, the results suggest that the observer makes errors of 4.6\% and 5.5\%, respectively. These values are very good for Soft Robotics applications. While they do not match the 2\% error margin achieved in~\cite{Bern2024}, this result is obtained at a lower computational cost with greater integration capacity with AR.

At first glance, studying the two axes separately may suggest that the errors are similar. However, the standard deviation of the Y-axis is higher than that of the X-axis. Similarly, the third quartile of errors on the Y-axis is half a millimetre lower than on the X-axis, meaning its estimates are more precise, but its outliers are larger.

Therefore, to improve the observer's accuracy, their estimates must be refined in the areas of greatest error. As mentioned, these areas coincide with the boundary of the workspace. This may be due to the linearity assumptions being less accurate the further the robot moves from its central position, or because dynamic effects occur in these areas, which are reached at higher speeds and are not taken into account by the observer. While the former issue can be resolved by applying a correction factor or by slightly relaxing the model's assumptions, albeit at the expense of increased computational complexity, the latter issue would necessitate a more in-depth study involving the application of more advanced simulation techniques.

\section{Conclusions}
\label{secc:concl}
This paper presents an augmented reality interface for the teleoperation of modular soft robots. While common in many areas of robotics, these interfaces are underdeveloped in soft robotics. This is primarily due to the difficulty of developing observers that are both fast and accurate.

This problem has been addressed in this article. Using Unity as the development software, a dual glasses-central computer system has been implemented. The glasses act as the interface, while the computer handles calculations and communication with the actual robot. When receiving data from the robot's sensors, the system uses its own physics, with some assumptions, to represent the robot's position. The results show errors of 5\% of the robot's length, very good accuracy compared to other developments in the literature, although performance could be improved in the boundary areas of the workspace. Furthermore, a future direction is proposed: validating the AR interface with real users to confirm its functionality to introduce improvements in its ease of use.


\section*{Acknowledgments}

This work is the result of research activities carried out at the Centre for Automation and Robotics, CAR (UPM-CSIC), in the facilities of the Escuela Técnica Superior de Ingenieros Industriales, within~the Robotics and Cybernetics research group (RobCib). This work is supported by “Ayudas para contratos predoctorales para la realización del doctorado con mención internacional en sus escuelas, facultad, centros e institutos de I+D+i”, which is funded by “Programa Propio I+D+i 2022 from Universidad Politécnica de Madrid” and~by the “Proyecto CollaborativE Search And Rescue robots (CESAR)” (PID2022-142129OB-I00) funded by MCIN/AEI/10.13039/501100011033 and “ERDF/EU”

\section*{Annex: Code}
The code of the two developed subsystems and for controlling the manipulator can be found in the following links:
\begin{itemize}
    \item \url{https://github.com/Robcib-GIT/PETER}
    \item \url{https://github.com/Robcib-GIT/PETER-H}
    \item \url{https://github.com/Robcib-GIT/PETER-DK}
\end{itemize}

\ifCLASSOPTIONcaptionsoff
  \newpage
\fi



\bibliographystyle{IEEEtran}
%

\bibliography{Bibliography/Mios, Bibliography/SoftRobotics, Bibliography/Nuevos}












\end{document}